\documentclass[letterpaper, 10 pt, conference]{ieeeconf}
\IEEEoverridecommandlockouts    

\usepackage{cite}
\usepackage{amsmath,amssymb,amsfonts}
\usepackage{algorithmic}
\usepackage{graphicx}
\usepackage{textcomp}
\usepackage{xcolor}
\usepackage{booktabs}
\usepackage{multirow}
\usepackage{hyperref}
\usepackage[table]{xcolor}
\usepackage{cuted}
\usepackage[font=normalfont,labelfont=bf]{caption}

\begin{document}

\title{Enhancing Multimodal Large Language Models \\ for Safety-Critical Driving Video Analysis}

\author{
 	\parbox{\textwidth}{%
 		\centering
 		Tomaso Trinci$^{1}$, Henrique Piñeiro Monteagudo$^{1}$, Leonardo Taccari$^{1}$%
 	}%
 	\thanks{$^{1}$Verizon Connect, Florence, Italy.}
 	\thanks{{\tt name.surname@verizonconnect.com}}
    }

\maketitle

\begin{strip}
  \begin{minipage}{\textwidth}
    \centering
    \vspace{-50pt} 
    \includegraphics[width=0.9\linewidth]{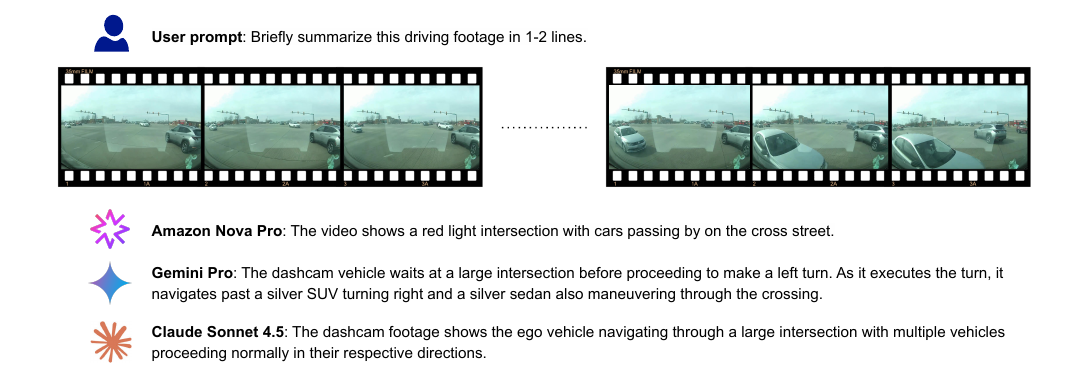}
    \captionsetup{type=figure}
    \captionof{figure}{\textbf{Current MLLMs struggle to identify Safety-Critical Events.} A qualitative failure case showing that state-of-the-art models fail to describe a clear collision. Even in a short (16s) sequence where the impact is the only significant event, no evaluated model successfully detects the crash.}
    \label{fig:teaser}
    \vspace{10pt} 
    \end{minipage}
\end{strip}

\begin{abstract}
Recent advancements in Multimodal Large Language Models (MLLMs) have demonstrated impressive capabilities in general visual understanding. However, their application to safety-critical driving scenarios remains limited by an inability to accurately perceive and reason about rare high-stakes dynamic events, such as collisions or near-collisions.
To address this, we introduce a pipeline that enhances MLLM perception by fusing downsampled video frames with synchronized high-frequency telematics data (IMU and GPS) and semantic insights from specialized computer vision models. 
Our pipeline generates high-quality pseudo-labels, including descriptive captions and question-answer pairs, specifically designed to train MLLMs to identify and describe Safety-Critical Events (SCEs) in real-world driving footage.
We show the effectiveness of our approach fine-tuning the open-source QwenVL-2.5 model via DoRA adapters: our experiments demonstrate significant improvements in identifying and explaining safety-critical events, with fewer than 50M trainable parameters and limited computational budget.
\end{abstract}

\section{Introduction}
The integration of Multimodal Large Language Models (MLLMs) into autonomous driving and ADAS applications promises a transformative shift towards systems capable of human-like reasoning and explainability. While current MLLMs excel at identifying static objects and generating high-level scene descriptions, they lack the ability to reason about complex dynamics and the causal chains of actions, frequently failing to capture the nuances of Safety-Critical Events (SCEs)~\cite{shi2025scvlm}. 
We categorize SCEs into three distinct levels of severity: \textit{collisions}, involving a physical impact between the ego vehicle and another entity; \textit{near-collisions}, characterized by emergency evasive actions such as hard braking; and \textit{normal driving}, representing baseline operation where the vehicle adheres to standard traffic flow without any anomalous maneuvers.
When presented with video footage of a collision or a near-collision event, state-of-the-art models frequently generate captions that describe the background scenery or unrelated traffic participants, while completely overlooking the accident itself.
Figure~\ref{fig:teaser} highlights the limitations of current state-of-the-art models, illustrating a case where general-purpose models fail to report a collision that is clearly visible from the footage.

We argue that this perceptual gap stems primarily from two factors. First, the scarcity of SCEs within public driving video datasets.
Secondly, the few existing public datasets with SCEs are small scale and typically offer only visual modalities, that alone are often ambiguous or insufficient for inferring latent physical forces, such as the abrupt deceleration characteristic of an impact. Conversely, other sensors, specifically Inertial Measurement Units (IMUs), can provide critical quantitative evidence regarding the vehicle's kinematic state.

To bridge this gap, we introduce a multimodal fusion pipeline that grounds visual representations using vehicle telemetry and domain-specific metadata from specialized ML models (e.g., traffic violation detectors~\cite{bravi2021detection, trinci2025color}). By integrating these explicit physical and semantic cues, we enable a state-of-the-art generalist MLLM to serve as a \textit{teacher}, generating high-fidelity, context-aware pseudo-labels for SCEs. We subsequently leverage these enriched descriptions to train a compact, domain-specific \textit{student} model. Through this cross-modal distillation process, the student model learns to reliably attend to critical events, achieving robust descriptive accuracy at inference time even in the absence of specialized metadata. Our main contributions are as follows:
\begin{itemize}
    \item A multimodal fusion pipeline that 1) grounds visual tokens with synchronized telemetry and semantic metadata from specialized models and 2) uses a state-of-the-art LLM as a teacher to generate high-quality pseudo-labels focused on the description of SCEs.
    \item Empirical evidence that fine-tuning a smaller student model enables high-quality SCE descriptions, even when IMU and semantic metadata are unavailable at inference time.
\end{itemize}

\section{Related Work}
By extending the capabilities of LLMs to process non-textual modalities like video and audio, MLLMs represent a huge advancement for Advanced Driver Assistance Systems and Autonomous Vehicles. Their ability to ground visual input in language is relevant for high-level tasks such as driving scene understanding and accident anticipation.

State-of-the-art proprietary model families, such as GPT, Claude, and Gemini, demonstrate the ability to ingest images and produce accurate scene descriptions. Recently, numerous open-source model families with similar capabilities have been released, including QwenVL~\cite{Qwen3-VL}, InternVL~\cite{wang2025internvl3}, Molmo~\cite{clark2026molmo2}, and MiniCPM-V 4.5~\cite{yu2025minicpm}. The architectural blueprints of these models are often similar: they use a vision encoder to process images and produce a latent representation. A fusion layer then converts this representation into visual tokens, which are concatenated with text tokens produced by the LLM tokenizer to incorporate textual prompts. These combined tokens are ingested by the LLM, which generates an answer autoregressively. 

Differences among these models arise primarily from their training methodologies, the specific pre-trained LLM used, and, crucially for driving applications, their approach to image preprocessing and positional encoding. For instance, the Qwen-VL family employs 3D Rotary Positional Embeddings~\cite{su2024roformer} to maintain both temporal and spatial consistency.

Despite numerous architectural refinements and the variety of available MLLMs, recent work demonstrates their inability to effectively perceive SCEs~\cite{shi2025scvlm, shi2025synshrp2}. This limitation arises from multiple factors. The impulsive nature of these events means critical dynamics are easily missed at the lower resolutions or frame rates often employed to drastically reduce token counts. Moreover, there is a lack of domain-specific data, especially from an egocentric perspective.

This scarcity becomes evident when examining the current landscape of driving datasets. While the field of autonomous driving has seen an explosion of datasets, such as LingoQA~\cite{marcu2024lingoqa}, DriveVLM~\cite{sima2024drivelm}, and BDD-X~\cite{kim2018textual}, that offer different types of ground truth to learn to reason on complex urban scenarios, these datasets fundamentally lack SCEs. A few smaller datasets, such as SHRP~\cite{hankey2016description}, DoTA~\cite{yao2022dota}, and Nexar~\cite{moura2025nexar}, focus specifically on SCEs; however, they often provide only simple event classifications, lacking the dense descriptions or Question-Answer pairs necessary to train MLLMs for this specific scenario. In this paper, we propose a pipeline capable of generating accurate video captions and QA pairs from real-world driving sequences containing SCEs. Our approach enables the effective fine-tuning of MLLMs in this domain.

\begin{figure*}[!tb]
  \centering
  \includegraphics[width=0.99\linewidth]{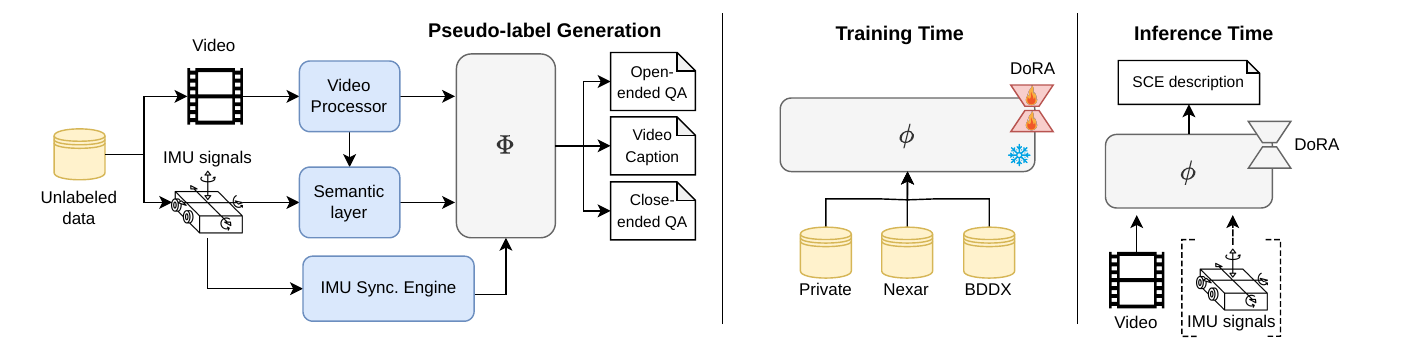}
  \caption{\textbf{High-level overview of our proposed solution.} The dataset is generated by ingesting unlabeled video and IMU data through a Video Processor and an IMU Synchronization Engine. These processed signals, combined with outputs from specialized models, serve as input to the teacher model $\Phi$ to generate captions alongside open-ended and closed-ended questions. These annotations are subsequently used to train adapters on a smaller student open-source model $\phi$, enabling the recognition of safety-critical events. Nexar and BDD-X datasets are utilized as supplementary data sources. During the inference phase, the adapters are merged with the base model, which processes the video input and IMU data (if available).}
  \label{fig:pipe_paper}
\end{figure*}

\section{Methodology}

The proposed framework operates on a multimodal input space consisting of videos and IMU telemetry. Let $\mathcal{V}=\{v_i\}_{i=1}^N$ denote a dataset consisting of $N$ video clips. Each video $v_i$ comprises footage captured by a windshield-mounted dashcam. The raw video clips have a duration of $16$ seconds, recorded at a frame rate of $30$ fps with a spatial resolution of $1280 \times 720$. These clips may contain safety-critical events. 
The corresponding IMU data for each video includes triaxial acceleration signals $\mathcal{A} = \{a_i\}_{i=1}^N$ and gyroscope signals $\Omega = \{\omega_i\}_{i=1}^N$, both sampled at $100$ Hz, alongside GPS speed $\mathcal{S}= \{s_i\}_{i=1}^N$ recorded at $1$ Hz.

A high-level overview can be seen in Figure~\ref{fig:pipe_paper}.
The following subsections detail the pipeline for multimodal alignment, the specialized semantic layer, the synthetic data generation process, and the subsequent distillation into a student model. To simplify notation in the subsequent sections, we consider a single arbitrary sample consisting of video $v$ and signals $a$, $\omega$, and $s$, dropping the index $i$.

\subsection{Video Processor}
\label{sec:video_processor}
The video processor ingests raw footage and identifies the timestamp of potential SCEs using a heuristic based on accelerometer data. We define the event timestamp, denoted as $t_e$, as the moment corresponding to the maximum absolute value of the g-sensor reading along the $x$-axis (indicative of peak acceleration or harsh braking). We use a temporal window spanning from $4$ seconds prior to the event to $2$ seconds after.
Within this window, frames are temporally downsampled to $3$ fps, yielding a sequence of $18$ frames.

\subsection{IMU Synchronization and Alignment}
\label{sec: imu_processing}
A critical component of our framework is the temporal alignment of high-frequency telematics data ($100$ Hz) with the downsampled video stream ($3$ fps). We introduce a synchronization mechanism that maps instantaneous sensor measurements to discrete video frames, ensuring strictly synchronized multimodal input.

\subsubsection{Accelerometer (g-sensor)}

We process the triaxial acceleration signals by partitioning the raw $100$ Hz data stream into non-overlapping windows corresponding to the duration of a single video frame. Given a raw sampling rate $f_s = 100$ Hz and a video framerate of $3$ fps, we define a window size of $W = \lfloor f_s / 3 \rfloor$ samples.
To align the data, we apply a block-averaging operation. Let $a_{\text{raw}}[n]$
denote the raw acceleration signal at index $n$. The synchronized acceleration value for the $k$-th video frame, denoted as $a_{\text{sync}}[k]$, is given by:
\begin{equation*}
a_{\text{sync}}[k] = \frac{1}{W} \sum_{n=(k-1)W+1}^{kW} a_{\text{raw}}[n].
\end{equation*}
This operation yields a sequence of $18$ acceleration values, temporally aligned with the  previously extracted video frames.

\subsubsection{Gyroscope}

For the gyroscope's $z$-axis, we compute the total angular displacement per frame. This allows the model to correlate visual changes in heading with the physical accumulation of rotation. We perform discrete time integration by summing the instantaneous rotation rates over the window $W$. The total angle change $\Delta\alpha_{\text{sync}}[k]$ for frame $k$ respect to frame $k-1$ is given by:
\begin{equation*}
\Delta\alpha_{\text{sync}}[k] = \sum_{n=(k-1)W+1}^{kW} \frac{\omega_{z}[n]}{f_s},
\end{equation*}
where $\omega_{z}[n]$ is the gyroscope measurement on the $z$-axis at index $n$.

\subsubsection{GPS Speed}
The GPS speed signal, originally sampled at $1$ Hz, requires upsampling to match the $3$ fps target frequency of the video and IMU inputs. We achieve this via linear interpolation.

\subsection{Specialized Vision \& Semantic Layer}
\label{sec:specialized_models}
To enhance the MLLM’s reasoning, we integrate a suite of domain-specific models. These models inject inductive biases related to traffic dynamics and safety rules, providing high-level semantic features to the MLLM.

\begin{itemize}
    \item \textbf{IMU-Based Crash Detection:} This module, inspired by~\cite{kubin2021deep, SHI2022106836}, ingests the sensors $a$, $\omega$, and $s$ which are processed through a series of 1D convolutional layers followed by RNN blocks to capture temporal dependencies. The model generates a binary classification indicating the occurrence of a collision event.
    
    \item \textbf{Harsh Driving Maneuver Classification}: This model ingests the same raw sensors signals together with raw video frames and bounding boxes from an object detector. These inputs are processed through convolutions and further fed to a Spatio-Temporal Attention Selector module and fully connected layer for classification across several harsh driving maneuver categories~\cite{simoncini2022unsafe}.
    
    \item \textbf{Traffic Light Violation:} Following the ideas in~\cite{polley2024tld, trinci2025color}, the model takes a video $v$ as input and solves a detection and classification task on multiple frames independently. It searches for traffic light sequences that indicate a violation, such as a continuous red light that never turns green before disappearing from the scene.
    
    \item \textbf{Stop Sign Violation:} This pipeline~\cite{bravi2021detection} receives sensor data and video as input. The video is processed by an image classifier to detect the possibility of a stop sign being present; these probabilities are fused with speed and acceleration to generate features which are fed to a downstream model which outputs a severity score.
\end{itemize}

We formally define $\mathcal{M}$ as the aggregate output set derived from these domain experts. $\mathcal{M}$ encompasses the specific event classifications described above, augmented by the continuous object tracking and bounding box coordinates generated by an object detector.

\subsection{Synthetic Supervision}
We adopt a knowledge distillation paradigm to transfer reasoning capabilities from a state-of-the-art \textit{teacher} MLLM $\Phi$ to a computationally efficient \textit{student} architecture $\phi$. The process begins with the generation of a synthetic dataset, $\mathcal{D}$, comprising detailed captions and Question-Answer (QA) pairs. To generate this supervision, the Teacher $\Phi$ ingests a multimodal input stream that includes the temporal sequence of 18 video frames, augmented with bounding box overlays, with the telemetry vectors ($a_{\text{sync}}, \Delta\alpha_{\text{sync}}, s_{\text{sync}}$) and the aggregate output $\mathcal{M}$ from the specialized semantic layer.

A \textit{System Prompt} instructs the model to describe the scene focusing on the SCE (if it exists) and formulate questions, while a \textit{User Prompt} interleaves image tokens with text tokens that encode physical sensor data and explicit object semantics (e.g., class, centroid, count). Formally, the target dataset $\mathcal{D}$ is sampled as:
\begin{equation*}
\mathcal{D} \sim \Phi(y | v, a_{\text{sync}}, \Delta\alpha_{\text{sync}}, s_{\text{sync}}, \mathcal{M})
\end{equation*}
This explicit conditioning ensures the generated targets are factually grounded in the scene's physical dynamics. Subsequently, the student model $\phi$ is fine-tuned to predict $\mathcal{D}$ using Weight-Decomposed Low-Rank Adaptation (DoRA)~\cite{liu2024dora}. 
The semantic metadata $\mathcal{M}$ is used to generate the targets, but the student is trained to generate descriptions with \textit{only} the raw video and (if available) telemetry. This enables the model to internalize the semantic insights of the specialized modules without requiring their execution during inference.
Finally, we also adopt an IMU \emph{drop-out} mechanism: we exclude IMU signal data in input with a 50\% probability, so that the model does not learn to rely solely on the IMU data and can also work in inference with video only.

\section{Experimental Results}
\label{sec:experimental-results}
\begin{figure}
  \centering
  \includegraphics[width=.99\linewidth]{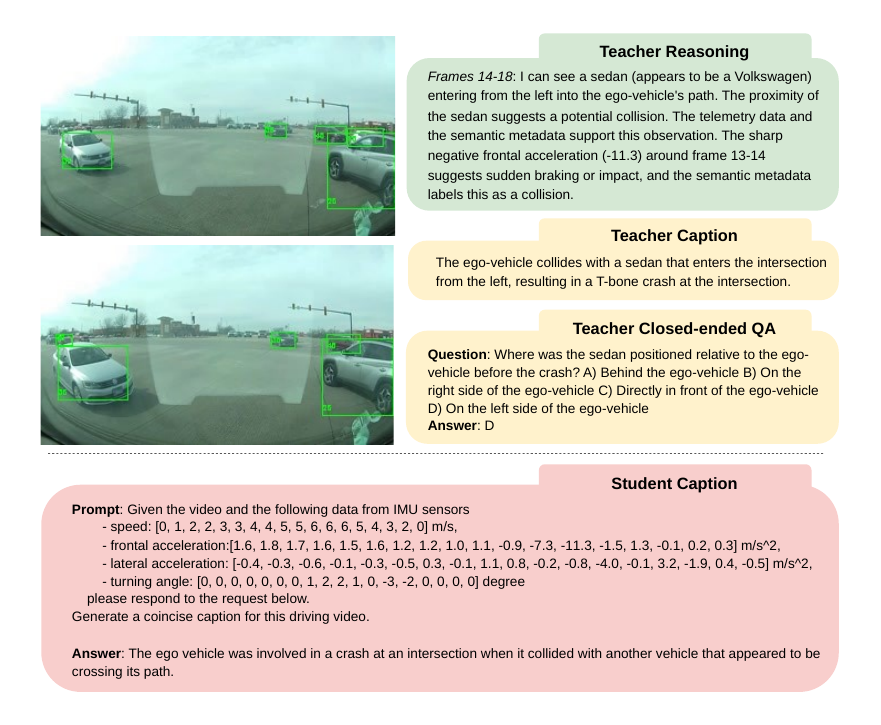}
  \caption{Pseudo-labels generated with our proposed pipeline, corresponding to the same event depicted in Fig.~\ref{fig:teaser}. The teacher successfully leverages the full spectrum of processed multimodal information, demonstrating the ability to correlate visual features (the appearance of the sedan) with IMU signals ($-11.3 m/s^2$ deceleration spike) to accurately reconstruct the scene's spatiotemporal dynamics. The student learns to correctly focus on the SCE.}
  \label{fig:reasoning}
\end{figure}

This section evaluates our fine-tuned model across various downstream tasks and benchmarks. 
Experiments utilized a private dataset of 7,000 US dashcam videos and telemetry, covering diverse lighting, weather, and camera mounting points.  Our pipeline generated over 80,000 high-quality pseudo-labels for fine-tuning.
Figure~\ref{fig:reasoning} presents a qualitative example of the generated pseudo-captions, demonstrating how the inclusion of IMU signals enable precise correlation between visual cues and vehicle dynamics.

\subsection{Experimental setting}
We employ Claude Sonnet as the teacher model and \textit{Qwen2.5-VL-7B-Instruct}~\cite{bai2025qwen2} as the student architecture. We apply DoRA~\cite{liu2024dora} to all linear layers of the student, while keeping both the vision encoder tower and the projection layer frozen. The adapters are configured with rank $r=32$ and scaling factor $\alpha=64$. Optimization is performed using AdamW with a learning rate of $5 \times 10^{-5}$. The model is trained with a batch size of $32$, with the image resolution fixed at $420 \times 240$ to limit the number of visual tokens. Each video clip spans a 6-second window centered around the event timestamp, $t_e$ (see Section~\ref{sec:video_processor}), and is sampled at 3 fps. To mitigate overfitting, we utilize NEFTune~\cite{jain2024neftune} regularization, setting the noise magnitude to 5.

To scale our dataset, we incorporate BDD-X~\cite{kim2018textual}, selecting 6-second clips with annotated actions and justifications. To improve SCE robustness, we also include 1,500 samples from the Nexar dataset~\cite{moura2025nexar}, trimmed around the moment of possible impact for binary classification (SCE vs. non-SCE). The final 18,000-video dataset follows a 90/5/5 split, with test sets segregated by source to allow separate evaluations of our private data and the Nexar benchmark.
Fine-tuning was completed in approximately 5 hours on four NVIDIA A10G GPUs. With fewer than 50M trainable parameters, the total infrastructure cost for the fine-tuning stage remained under \$100. 

\subsection{Evaluation on Downstream Tasks}
We compare our adapted model against six baselines. These include two state-of-the-art proprietary models, Amazon Nova Pro and Claude Sonnet 4.5, as well as four recent open-source models, Molmo 2~\cite{clark2026molmo2}, MiniCPM-V 4.5~\cite{yu2025minicpm}  InternVL 3.5~\cite{wang2025internvl3} and QwenVL 3.5~\cite{Qwen3-VL}. 
Additionally, we evaluate the base \textit{QwenVL-2.5} model in a zero-shot setting to establish a baseline before adaptation. We assess performance across three distinct tasks: \textit{(i) Captioning}, evaluated via ROUGE-L F1 and BERTScore F1; \textit{(ii) Closed QA}, evaluated via accuracy on static scene questions; and \textit{(iii) SCE Classification}, evaluated via accuracy, which requires discriminating between normal driving, near-collision, and collision events. All reported metrics are evaluated on the held-out test split.

\subsubsection{Scene Understanding in Safety-Critical Event}

Table~\ref{tab:vzc-sce-full} presents the results on the test set of the pseudo-labeled dataset. In this evaluation, \textit{all} models receive IMU data as supplementary context within the prompt.
The results indicate a performance plateau in captioning metrics (ROUGE-L and BERTScore) across both large proprietary systems and smaller open-source models. This phenomenon stems from the nature of our ground truth captions that focus on the description of the SCEs, detailing how they occur and which vehicles are involved. Standard models tend to generate static, generic descriptions, as illustrated qualitatively in Figure~\ref{fig:teaser}. While these generic captions achieve moderate scores, they fail to capture SCEs. In contrast, our DoRA-adapted model demonstrates a significant improvement, achieving a ROUGE-L of 0.44 and BERTScore of 0.51, confirming its ability to align with the dense, event-centric ground truth. In the Closed QA task, we observe generally strong results across all models, as the questions also pertain to static objects within the scene. In particular, large commercial models such as Sonnet perform exceptionally well on this task. However, our fine-tuned model successfully bridges the performance gap between small open-source models and large commercial baselines. Finally, accuracy on the three-class SCE classification task serves as a robust and more objective metric for scene understanding. To ensure a consistent assessment for this task, the test set labels for event classification were human-verified. While general-purpose models fail to perform effectively on this task, our specialized model demonstrates significantly higher accuracy. We acknowledge that for pure video classification, models like the V-JEPA~\cite{assran2025vjepa2selfsupervisedvideo} family are likely a superior solution if properly trained. However, we employ this task here specifically as a proxy to evaluate scene understanding, as our primary objective remains developing a model capable of accurately describing the scene even in the presence of out-of-distribution events such as a collision.

\begin{table}
    \centering
    \caption{Evaluation on our private dataset (video + IMU input). We report ROUGE-L and BERTScore for captioning, and accuracy for Closed QA and SCE 3-class problem (normal, near-collision, collision).}
    \label{tab:vzc-sce-full}
    \resizebox{0.48\textwidth}{!}{%
    \begin{tabular}{l r c c c c}
        \toprule
        \textbf{Model} & \textbf{Size} & \textbf{ROUGE-L} & \textbf{BERTScore} & \textbf{Closed QA} & \textbf{SCE} \\
         \cmidrule(lr){1-1}
          \cmidrule(lr){2-2}
          \cmidrule(lr){3-4}
           \cmidrule(lr){5-5}
            \cmidrule(lr){6-6}
          Nova Pro    & - & 0.17 & 0.19 & 87.5 & 53.8 \\
          Sonnet 4.5    & - & 0.18 & 0.17 & \textbf{93.4} & 53.9 \\
          Molmo 2 & 8B & \underline{0.19} & 0.19 & 85.5 & \underline{59.2} \\
          InternVL 3.5  & 8B & 0.15 & 0.19 & 90.8 & 49.2 \\
          MiniCPM-V 4.5  & 8B & 0.18 & \underline{0.22} & 88.1 & 49.2 \\
          QwenVL 3.5    & 8B & 0.18 & 0.18 & 88.5 & 56.9 \\
          \cmidrule(lr){1-1}
          \cmidrule(lr){2-2}
          \cmidrule(lr){3-4}
           \cmidrule(lr){5-5}
            \cmidrule(lr){6-6}
          QwenVL-2.5    & 7B & 0.18 & 0.16 & 83.2 & 46.1 \\
          -- DoRA Adapted (Ours)      & 7B  & \textbf{0.44} & \textbf{0.51} & \underline{92.7} & \textbf{87.7} \\
        \bottomrule
    \end{tabular}
    }
\end{table}

Table~\ref{tab:binary_task} presents the results for the binary SCE identification task, where we group near-collision and collision events into a single ``positive" class.
This setting enables the testing of our models on the Nexar dataset, which features human-verified labels despite lacking explicit near-collision annotations.
The results suggest that even in this simplified formulation, generalist models underperform on both datasets. We can observe that, while they are fairly precise in their predictions, they exhibit poor recall. In contrast, our fine-tuned model achieves a superior trade-off between precision and recall. 
On our internal data, the adapted model obtains significantly better results compared to Nexar. We attribute this gap to two reasons: \textit{first}, the training data distribution, as the Nexar domain constitutes only 5\% of the total training set; \textit{second}, the absence of IMU data to enrich the prompt context (see Sec. \ref{imu-ablation} for more results in absence of IMU).

\begin{table}
    \centering
    \caption{Evaluation on our private dataset (video + IMU input) and on Nexar (video only) for the binary SCE identification problem, collision/near-collision vs normal. We report precision, recall, and accuracy.}
    \label{tab:binary_task}
    \resizebox{0.48\textwidth}{!}{%
    \begin{tabular}{l l r cc c}
        \toprule
         & \textbf{Model} & \textbf{Size} & \textbf{Prec. ($y=1$)} & \textbf{Rec. ($y=1$)} & \textbf{Acc.}\\
          \cmidrule(lr){2-2}
          \cmidrule(lr){3-3}
          \cmidrule(lr){4-5}
          \cmidrule(lr){6-6}
        
        \multirow{8}{*}{\rotatebox[origin=c]{90}{\textbf{Private}}}
          & Nova Pro    & -  & \textbf{96.1} & 35.2 & 63.8 \\
          & Sonnet 4.5   & - & 81.8 & 63.3 & \underline{72.3} \\
          & Molmo 2   & 8B & 76.5 & \underline{69.0} & 71.5 \\
          & InternVL 3.5  & 8B  & 81.8 & 25.3 & 56.1 \\
          & MiniCPM-V 4.5  & 8B  & 91.6 & 31.0 & 60.8 \\
          & QwenVL 3.5  & 8B & 81.6 & 63.3 & 72.2\\
          \cmidrule(lr){2-2}
          \cmidrule(lr){3-3}
          \cmidrule(lr){4-5}
          \cmidrule(lr){6-6}
          & QwenVL 2.5    & 7B  & 72.7 & 45.0 & 60.7 \\
          & -- DoRA Adapted (Ours)      & 7B & \underline{93.3} & \textbf{87.3} & \textbf{90.0 }\\
        \midrule
        
        \multirow{8}{*}{\rotatebox[origin=c]{90}{\textbf{Nexar}}}
          & Nova Pro    & -  & 64.0 & 21.1 & 54.3 \\
          & Sonnet 4.5   & - & 75.0 & 35.5 & \underline{61.5} \\
          & Molmo 2   & 8B  & 58.8 & \textbf{92.1} & 63.6 \\
          & InternVL 3.5   & 8B  & \textbf{85.7} & \phantom{0}7.9 & 52.9 \\
          & MiniCPM-V 4.5  & 8B & 58.3 & 18.4 & 52.3 \\
          & QwenVL 3.5  & 8B & 79.2 & 25.0 & 58.9 \\
          \cmidrule(lr){2-2}
          \cmidrule(lr){3-3}
          \cmidrule(lr){4-5}
          \cmidrule(lr){6-6}
          & QwenVL 2.5    & 7B  & 77.7 & 27.6 & 59.6 \\
          & -- DoRA Adapted (Ours)    & 7B  & \underline{85.6} & \underline{63.1} & \textbf{76.1} \\
        
        \bottomrule
    \end{tabular}
    }
\end{table}

\subsubsection{Scene Understanding in General Driving Scenarios}
Table~\ref{tab:lingoqa} reports the evaluation metrics on the LingoQA dataset. In contrast to safety-critical benchmarks, LingoQA emphasizes common, everyday driving scenarios. We leverage this evaluation to verify that our fine-tuned model does not overfit to its specialized domain at the expense of general driving knowledge. 
Unsurprisingly, general-purpose models exhibit strong performance on this task, given that regular, uneventful driving scenes align more closely with the broader data distributions seen during their training. It is worth noting that our fine-tuned model demonstrates substantial improvement over its off-the-shelf counterpart. This shows that our training process, designed for specialization in safety-critical events, is able to preserve (and even improve) core semantic understanding of generic driving scenes.

\begin{table}
    \centering
    \caption{Evaluation on the LingoQA dataset. We report ROUGE-L and BERTScore for open-ended QA task.}
    \label{tab:lingoqa}
    \resizebox{0.4\textwidth}{!}{%
    \begin{tabular}{l r c c}
        \toprule
        \textbf{Model} & \textbf{Size} & \textbf{ROUGE-L} & \textbf{BERTScore} \\
          \cmidrule(lr){1-1}
          \cmidrule(lr){2-2}
          \cmidrule(lr){3-4}
          Nova Pro    & - & 0.25 & \textbf{0.38}  \\
          Sonnet 4.5  & -  & 0.16 & 0.23   \\
          Molmo 2 & 8B & 0.19 & 0.27  \\
          InternVL 3.5  & 8B & 0.25 & 0.33 \\
          MiniCPM-V 4.5  & 8B & 0.23 & \underline{0.35}  \\
          QwenVL 3.5    & 8B & \textbf{0.32} & 0.32  \\
          \cmidrule(lr){1-1}
          \cmidrule(lr){2-2}
          \cmidrule(lr){3-4}
          QwenVL-2.5    & 7B & 0.25 & 0.24  \\
          -- DoRA Adapted (Ours)     & 7B & \underline{0.28} & 0.31  \\
        \bottomrule
    \end{tabular}
    }
\end{table}

\subsection{Ablation Studies}
\subsubsection{Impact of Model Scale and DoRA Rank}
We evaluate the impact of DoRA rank dimensions and compare the performance relative to a smaller base model.
Table~\ref{tab:dora_size_abl} reports the metrics obtained on our internal dataset. 
As expected, the performance of the 3B model is inferior to that of the 7B variant; however, increasing the DoRA rank helps bridge the performance gap with the larger model. 
Conversely, we observe that this trend does not hold for the 7B model, where relatively small adapters yield superior results compared to larger ones, particularly in classification. 
We hypothesize that the dataset lacks the scale to prevent overfitting with an excessive number of trainable parameters.

\begin{table}
    \centering
    \caption{Ablation study of DoRA rank across model scales. The selected configuration is highlighted in blue.}
    \label{tab:dora_size_abl}
    \resizebox{0.40\textwidth}{!}{%
    \begin{tabular}{l r c c c c}
        \toprule
        \textbf{Size} & \textbf{Rank} & \textbf{ROUGE-L} & \textbf{BERTScore} & \textbf{Closed QA} & \textbf{SCE} \\
         \cmidrule(lr){1-1}
          \cmidrule(lr){2-2}
          \cmidrule(lr){3-4}
           \cmidrule(lr){5-5}
            \cmidrule(lr){6-6}
          3B    & $8$ & 0.41 & 0.48 & 91.0 & 76.2 \\
          3B    & $16$ & 0.42 & 0.49 & 92.0 & 79.9 \\
          3B    & $32$ & 0.42 & 0.50 & 92.1 & 83.8 \\
          \cmidrule(lr){1-1}
          \cmidrule(lr){2-2}
          \cmidrule(lr){3-4}
           \cmidrule(lr){5-5}
            \cmidrule(lr){6-6}
          7B    & $8$ & 0.43 & 0.51 & 91.5 & 81.4 \\
          7B    & $16$ & 0.43 & 0.51 & 92.1 & 88.5 \\
          \rowcolor{blue!20}
          7B    & $32$ & 0.44 & 0.51 & 92.7 & 87.7 \\
          7B    & $64$ & 0.43 & 0.51 & 92.1 & 84.6 \\
        \bottomrule
    \end{tabular}
    }
\end{table}

\subsubsection{Impact of the IMU Modality}
\label{imu-ablation}
Table~\ref{tab:vzc-sce-wo-imu} reports the results for the closed QA and classification tasks when the input excludes IMU measurements. Interestingly, all models show a decrease in performance across every task, suggesting they capture semantic information from the IMU data even when not explicitly trained on it. Regarding the fine-tuned model, performance degrades but remains significantly higher than the zero-shot baselines, indicating that the IMU drop-out mechanism is effective in training a model that can reason even without telemetry.
We observe the most substantial drop in the three-class problem, where telematics appears particularly critical for distinguishing between near-misses from collisions, and normal driving. Additionally, our two-class results on the private dataset now align with the Nexar results, which is expected since the Nexar data lacks this telemetry data.

\begin{table}
    \centering
    \caption{Ablation study on our private dataset with video only (IMU signals were excluded from the input prompt). Values in parentheses denote the change in performance compared to experiments incorporating IMU data.}
    \label{tab:vzc-sce-wo-imu}
    \resizebox{0.48\textwidth}{!}{%
    \begin{tabular}{l r c c c}
        \toprule
        \textbf{Model} & \textbf{Size} & \textbf{Closed QA} & \textbf{SCE (3-cls.)} & \textbf{SCE (2-cls.)} \\
         \cmidrule(lr){1-1}
          \cmidrule(lr){2-2}
          \cmidrule(lr){3-3}
           \cmidrule(lr){4-4}
           \cmidrule(lr){5-5}
          Nova Pro    & - & 85.0 {\tiny(-2.5)} & 54.6 {\tiny (+0.8)} & 59.5 {\tiny (-4.3)}  \\
          Sonnet 4.5    & - & \textbf{92.1} {\tiny(-1.3)} & 53.7 {\tiny (-0.2)} & 56.7 {\tiny(-15.6)} \\
          Molmo 2 & 8B & 83.1 {\tiny(-2.4)} & \underline{55.3} {\tiny(-3.9)} & \underline{68.8} {\tiny(-2.7)} \\
          InternVL 3.5  & 8B & 86.0 {\tiny(-4.8)} & 39.7 {\tiny(-9.5)} & 46.1 {\tiny(-10.0)} \\
          MiniCPM-V 4.5  & 8B & 81.8 {\tiny(-6.3)} & 41.8 {\tiny(-7.4)} & 50.4 {\tiny(-10.4)}  \\
          QwenVL 3.5    & 8B & 84.8 {\tiny(-3.7)} & 47.5 {\tiny(-9.4)} & 53.1 {\tiny(-19.1)} \\
          \cmidrule(lr){1-1}
          \cmidrule(lr){2-2}
          \cmidrule(lr){3-3}
           \cmidrule(lr){4-4}
           \cmidrule(lr){5-5}
          QwenVL-2.5    & 7B  & 78.8 {\tiny(-4.4)} & 44.7 {\tiny(-1.4)} & 59.6 {\tiny(-1.1)}\\
          -- DoRA Adapted (Ours)      & 7B  & \underline{92.0} {\tiny(-0.7)} & \textbf{74.5} {\tiny(-13.2)} & \textbf{80.9} {\tiny(-9.1)}\\
        \bottomrule
    \end{tabular}
    }
\end{table}

\section{Conclusion and Limitations}
\label{sec:conclusion}
In this work, we address the inability of current MLLMs to perceive SCEs in driving footage. We introduce a multimodal pipeline that uses synchronized telematics and semantic metadata to generate dense pseudo-labels. Using this approach to fine-tune Qwen2.5-VL, we demonstrate that we significantly outperform open and commercial  baselines in detecting and describing SCEs.

Our training pipeline relies on synchronized video and IMU data, preventing its applicability to uninstrumented dashcam footage. Additionally, our current method encodes sensor inputs as simple textual sequences. Future work will explore integrating a dedicated \textit{time-series encoder} to better project high-frequency IMU data directly into the MLLM embedding space enhancing physical reasoning capabilities.

\bibliographystyle{IEEEtran}
\bibliography{references}

\end{document}